\documentclass[letterpaper, 10pt, conference]{ieeeconf}
\IEEEoverridecommandlockouts
\usepackage{graphicx,epstopdf,multirow,multicol,subfigure,booktabs,pgfplots}
\usepackage{amsmath}
\usepackage{float}
\usepackage{bm}
\usepackage{cite}
\usepackage{breqn}
\usepackage{amssymb,amsthm}
\usepackage{algorithm2e}
\usepackage{times}
\usepackage[T1]{fontenc}
\usepackage[colorlinks]{hyperref}

\usepackage{booktabs}
\allowdisplaybreaks
\newcommand\textlcsc[1]{\textsc{\MakeLowercase{#1}}}

\title{\LARGE \bf
Routing Unmanned Vehicles in GPS-Denied Environments
}

\author{Kaarthik Sundar$^{\dagger}$\thanks{$^{\dagger}$ Center for Non-Linear Studies, Los Alamos National Laboratory, NM. \texttt{kaarthik01sundar@gmail.com}},\;
Sohum Misra$^{*}$\thanks{$^{*}$ Dept. of Aerospace Engg., University of Cincinnati, OH. \texttt{}},\;
Sivakumar Rathinam$^{\ddagger}$\thanks{$^{\ddagger}$ Dept. of Mechanical Engg., Texas A\&M University,
College Station, TX.},\;
Rajnikant Sharma$^{*}$} \;

\renewcommand{\baselinestretch}{1.15}

\begin{document}

\maketitle
\thispagestyle{empty}
\pagestyle{empty}

\begin{abstract}
Most of the routing algorithms for unmanned vehicles, that arise in data gathering and monitoring applications in the literature, rely on the Global Positioning System (GPS) information for localization. However, disruption of GPS signals either intentionally or unintentionally could potentially render these algorithms not applicable. In this article, we present a novel method to address this difficulty by combining methods from cooperative localization and routing. In particular, the article formulates a fundamental combinatorial optimization problem to plan routes for an unmanned vehicle in a GPS-restricted environment while enabling localization for the vehicle. We also develop algorithms to compute optimal paths for the vehicle using the proposed formulation. Extensive simulation results are also presented to corroborate the effectiveness and performance of the proposed formulation and algorithms.
\end{abstract}


\section{Introduction} \label{sec:intro}
Unmanned Vehicles (UVs), both aerial and ground, have been finding its use in a plethora of civilian \cite{Johnson2017,Ou2014} and indoor applications (see \cite{Nonami2007,Chen2014,Tomic2012} and references therein). Knowledge of position and orientation is necessary to ensure decision-making and autonomy for these vehicles. The problem of localization deals with the estimation of position and orientation of the UVs. Localization therefore requires sensing, and correspondingly, localization procedures are dependent of the available sensory measurements. Since sensory measurements are usually contaminated with noise, the problem of localization also requires filtering the noise in order to determine an accurate estimate of location and orientation. The investment in GPS-based vehicle positioning systems that rely on the GPS measurements have garnered a lot of attention in the literature \cite{Reina2007}. However, most indoor environments and many parts of the urban canyon do not have access to GPS; even if available, the access is intermittent and not reliable. Hence, localization in a GPS-denied or GPS-restricted environment is an active area of research; it also has additional advantages of robustness, efficiency, and flexibility \cite{Sharma2013}.

In this paper, we present a joint route optimization and localization algorithm in a GPS-denied environment to visit a set of targets while enabling localization as the UV traverses its route. The proposed approach uses known objects in the environment referred to as landmarks (LMs) to aid in localization. The UV determines it relative position and orientation with respect to the LMs using exteroceptive sensor like camera, laser, etc. and assess its motion and attitude information such as velocity and turn rate using interoceptive sensors like IMU, encoders, etc. This enables the UV to estimate its states and localize itself in a GPS-denied environment. Conditions under which the UV would be able to estimate its states and localize itself using the LMs are first derived. These conditions are then embedded into an optimization framework where the sequence in which the UV should visit the targets, the route it takes, and the location where the LMs should be placed to enable localization are optimized. In this way, we are ensured that the UV can perform localization using the LMs as it traverses its route and visits the targets. The problem statement is can be summarized as follows:

\noindent \textit{Given a UV stationed at a depot, a set of target locations, and a set of potential locations where LMs can be placed, find a route for the UV and a subset of potential LM locations where LMs are placed such that the following conditions are satisfied:
\begin{enumerate} 
\item the route starts and ends at the depot and visits every target at least once, 
\item the UV should be able to estimate its position and orientation from the LMs as it traverses its route, and
\item the sum of the total travel distance and the number of LMs used is a minimum.
\end{enumerate}
}
We remark that, it is also easier to think about the above problem as a single vehicle routing problem with additional constraints for LM placement to aid in localization. Henceforth, we shall refer to this problem as a single vehicle routing problem using LMs to aid localization, abbreviated as SVRP-LL. SVRP-LL, being a generalization of the traveling salesman problem (TSP), is NP-hard. 

\section{Related work} \label{sec:lit_review}
The problem of localization of a vehicle, aerial and ground, in urban and indoor environments where GPS signals are not always reliable is well studied in the literature. In particular, authors in \cite{Wong2014} used techniques in computer vision to address the problem. Numerous variants of Simultaneous Localization and Mapping (SLAM) techniques have also been developed for high precision vehicle localization \cite{Levinson2007,Weiss2011}. Another frequently used method is infrastructure-aided localization for aerial and ground vehicles. In particular, infrastructure capable of providing range measurements to the vehicles are pre-installed in the environment and algorithms are developed to estimate the position and orientation of the vehicles \cite{Mao2007}. 

The problem of localization is also of independent interest to infrastructure-aided localization for transportation applications. The idea of infrastructure-aided navigation and control for automated vehicles is not new and has been considered at least since California PATH's Automated Highway Systems (AHS) program. However, this idea is useful for other applications such as Advanced Traffic Management Systems (ATMS) and Advanced Traveler Information System (ATIS); one can design V2I \cite{Doan2009} (vehicle to infrastructure) communication protocols by which the type of vehicle is identified along with the time stamp for the communication thereby obviating the need for traffic measurement devices such as loop detectors which are error-prone. Authors in \cite{Ou2014} proposed a Vehicular Ad-Hoc Network (VANET) based localization approach in which each vehicle estimates its location on the basis of beacon messages broadcast periodically by pairs of Road Side Units (RSUs) deployed on either side of the road. Authors in \cite{Khattab2015} modified the previous approach by using two-way time of arrival information to localize vehicles based on communication with a single road side unit (RSU or LM) instead of having 2 RSUs/LMs.

The first work to consider placement of LMs given a paths for multiple vehicles is \cite{Rathinam2015}. The authors formulated the landmark placement problem as a multiple vehicle path covering problem and presented an approximation algorithm using geometric arguments to address the problem.  This article is an extension of the work in \cite{Rathinam2015} on three fronts: (1) we formulate the joint vehicle routing and landmark placement problem as a mixed-integer linear program (MILP), (2) we present an algorithm to compute an optimal solution to the MILP, and finally, (3) we present extensive computational and simulation results showing the effectiveness of the proposed approach with a suitable estimation algorithm. The rest of the paper is organized as follows: in Sec. \ref{sec:vm}, the vehicle model and the algorithm to localize the vehicle when the position of the LMs are known a priori are detailed; conditions under which the localization algorithm would provide accurate position and orientation estimates are also discussed. Sec. \ref{sec:defn}, \ref{sec:formulation}, and \ref{sec:bandc} present the optimization problem definition, formulation, and the branch-and-cut algorithm to solve the SVRP-LL to optimality, respectively. Finally, the controller architecture and the computational results are detailed in Sec. \ref{sec:arch} and \ref{sec:results}, respectively.

\section{Vehicle model \& Localization algorithm} \label{sec:vm}
For the localization problem, an unmanned ground vehicle or an unmanned aerial vehicle flying at a constant altitude and traveling with a constant velocity $v$ is considered; the kinematic constraints of the vehicle in state-space form is as follows:

\begin{flalign}
    \dot{\bm x} = f(\bm x, \bm u) \triangleq
    \begin{bmatrix}
        v \cos(\psi)\\
        v \sin(\psi)\\
        \omega
    \end{bmatrix}
\end{flalign}

where, $v$ is the velocity of the vehicle, $\psi$ is the heading of the vehicle and $\omega$ is the rate of change of heading with respect to time. The vector $\bm x$ is the vector of state variables given by $(x, y, \psi)^T$; here $x$ and $y$ denote the position of the vehicle and $\psi$ denotes the heading information. The control input vector, $\bm u$, for the vehicle consists of $v$ and $\omega$. The interoceptive sensors detect the velocity, $v$, and angular velocity, $\omega$, of the vehicle. The exteroceptive sensors are used to detect the relative bearing measurement of the vehicle with respect to the known LMs. The vehicle is assumed to have a complete $360^{\circ}$ field of view. Without loss of generality, it is considered that the vehicle cannot move backwards, \emph{i.e.}, $v \geqslant 0$. Furthermore, the sensing range of the vehicle's exteroceptive sensor, denoted by $\rho_s$, is assumed to be constant. 


To route the UVs, the precise knowledge of position and heading of the vehicles is necessary. In GPS-denied environments, relative position measurements using range or bearing angle sensors to known landmarks can be used to localize the vehicle. An Extended Kalman Filter (EKF) or its information form given by the Extended Information Filter (EIF) can be used to estimate the vehicle's states $\bm x$. This article uses the EIF instead of the EKF for to estimate the vehicle's state $\bm x$ as the EIF is more favorable from a computational perspective. The estimation algorithm will provide meaningful localization estimates (consistent and bounded) if and only if the sensors provide enough information for localization, or in other words, if the system is observable. It has been shown that the bound on uncertainty is related with the eigenvalues of the observability gramian \cite{Song1992}. In our previous work \cite{Sharma2012}, we have shown that the UV needs bearing angle measurements from 2 different landmarks in order for the system to be observable using the Hermann-Krener criterion \cite{Hermann1977} for local observability. This technique of state estimation using the bearing information alone is referred to as \textit{bearing-only localization} \cite{Sharma2012, Sharma2013}.

\begin{figure}
		\centering
		\includegraphics[width = 60mm,keepaspectratio]{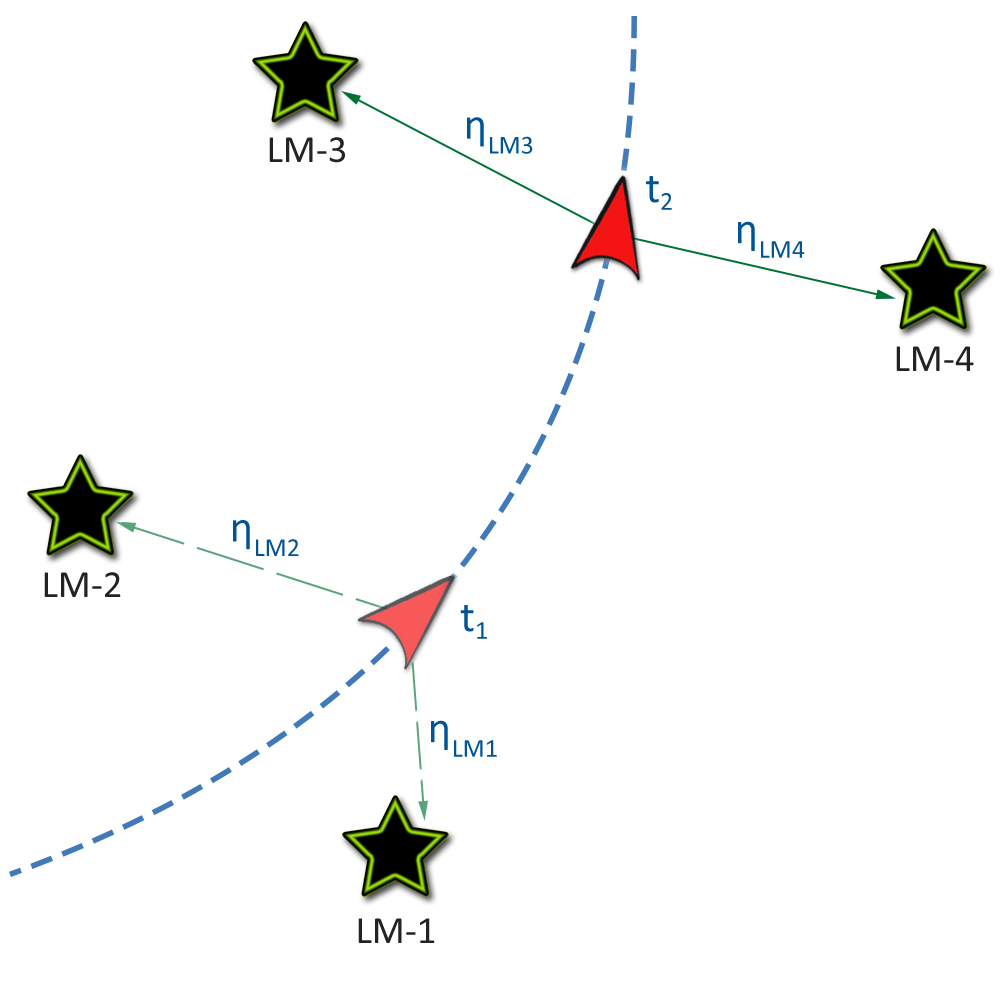}
		\caption{Relative Position Measurement Graph (RPMG) illustrating the conditions for enabling convergence of the estimation algorithm to the states of the system. The stars represent LMs and the red arrows represent the vehicle at times $t_1$ and $t_2$, respectively. The edge $\eta_{LM_i}$, $i=1,\dots,4$ indicates that the vehicle receives the bearing information from the landmark $LM_i$.}
		\label{fig:RPMG}
\end{figure}

The condition for enabling accurate estimation of the states of the vehicle can also be illustrated using a relative position measurement graph (RPMG). A RPMG is a graph that is used to represent the interaction and information flow between the vehicle and LMs. The RPMG consists of two types of nodes: the vehicle position at different time instants and the landmarks. An edge between a landmark $i$ and the vehicle at a particular time instant $t$ indicates that the vehicle obtains bearing measurement from the landmark $i$. An example of RPMG with single vehicle and multiple LMs with edges between the vehicle at time $t_1$ and $t_2$ and the landmarks is illustrated in Fig. \ref{fig:RPMG}. The main result of \cite{Sharma2012} that will be used in the forthcoming sections is that for the estimation algorithm, using an EIF, to provide accurate localization estimates for the vehicle at any given time $t$, the RPMG should contain at least two edges from the node that represents the vehicle at time $t$ to two different LMs. We also remark that having path to more than 2 LMs provides more information to the vehicles, thereby quickening the convergence rate of the estimation algorithm. 

\section{Optimization problem definition}  \label{sec:defn}
In the previous section, we elucidated the condition that enables localization of the vehicle from bearing measurements which is that, at any given instant of time the vehicle requires bearing information from at least two LMs for observability and for the estimation algorithm to converge to accurate state estimates. We shall now enforce this condition explicitly in the optimization problem to jointly optimize the routes and landmark placement. To that end, let $V$ denote the set of targets $\{v_1, \dots, v_n\}$ and let the vehicle be initially stationed at $v_1$. Let $K$ denote the set of potential locations where a LM can be placed. Associated with each vertex $v_i$ is a subset of locations, $K_{v_i}$; if a LM is placed in a location in $K_{v_i}$, then it can provide bearing measurement for the vehicle when the vehicle is at the target $v_i$. We note that since the vehicle has a $360^{\circ}$ field of view and the sensing range of the exteroceptive sensor on the vehicle is $\rho_s$, a location $k$ is in the set $K_{v_i}$ if and only if the distance between the location and the target $v_i$ is less than $\rho_s$, the sensing range of the exteroceptive sensor. The vehicle can perform localization along an edge $e \in \{(v_i,v_j): i<j\}$ if it has bearing measurements from at least 2 LMs as it traverses that edge. Hence, associated with each edge $e \in \{(v_i,v_j): i<j\}$ is a subset of potential LM locations, $K_e \subseteq K$; for a given edge $e \in \{(v_i,v_j): i<j\}$, $k \in K_e$ if and only if $k \in K_{v_i} \cap K_{v_j}$. For the vehicle to be able to localize itself as it traverses the edge $e$, LMs have to placed at at least two locations in $K_e$. Now, the problem is defined on an undirected graph $G=(V, E)$ where $E$ is the set of edges $\{e=(v_i, v_j):i<j\}$. We say that a set of LMs ``cover'' an edge $e$ if they are installed in a subset of locations $K_1$ such that $|K_1 \cap K_e| \geq 2$. The set of LMs that cover an edge can provide bearing measurements for the vehicle as it traverses that edge and thereby enabling the vehicle localization. Each edge $(v_i, v_j) \in E$ is associated with a non-negative cost $c_{ij}$ required to travel from target $i$ to target $j$. Also associated with each location $k\in K$ is a non-negative cost $d_k$ that denotes the cost for installing a LM at location $k$; if we wish to minimize the number of LMs placed then each $d_k$ takes a value $1$. The SVRP-LL consists for finding a path for the vehicle and placing/installing LMs in a subset of locations $K$ such that (i) each target in the set $V$ is visited at least once by the vehicle, (ii) each edge traversed by the vehicle is covered by at least two installed LMs , and (iii) the sum of the cost of the path and installation cost of the LMs is a minimum.

\section{Mathematical formulation} \label{sec:formulation}
We now present a mathematical formulation for the SVRP-LL, inspired by the models for the standard routing and covering problems \cite{Toth2014} (see \cite{Sundar2016,Sundar2015,Manyam2016,Sundar2016a,Sundar2016b,Sundar2016c} for routing and path planning problems concerning UVs). We associate to each feasible solution $\mathcal F$ for the SVRP-LL, a vector $\bm x \in \mathbb{R}^{|E|}$ (a real vector indexed by the elements of the edge set $E$) where each component $x_e$ of $\bm x$ takes a value one if the vehicle traverses the edge and zero otherwise. Similarly, associated to $\mathcal F$, is also a vector $\bm y \in \mathbb{R}^{|K|}$; the value of the component $y_k$ associated with a potential sensor location $k \in K$ is equal to one if a sensor is placed in the location $k$ and zero otherwise.

For any $S \subseteq V$, we define $\delta(S) = \{(i,j) \in E: i\in S, j\notin S\}$. If $S = \{i\}$, we simply write $\delta(i)$ instead of $\delta(\{i\})$. Finally for any $\mathcal E \subset E$, we define $x(\mathcal E) = \sum_{e \in \mathcal E} x_e$. Using the notations introduced thus far, the SVRP-LL is formulated as a mixed-integer linear program as follows:
\begin{flalign}
& \min \sum_{e\in E} c_e x_e + \sum_{k\in K} d_k y_k &\label{eq:obj} &\\
& \text{subject to: } \notag &\\
& x(\delta(v_i)) = 2 \quad \forall v_i \in V, &\label{eq:degree} \\
& x(\delta(S)) \geqslant 2 \quad \forall S \subset V, &\label{eq:sec} \\
& \sum_{k \in K_e} y_k \geqslant 2 x_e \quad\forall e \in E, \text{ and } &\label{eq:sensor} \\
& x_e \in \{0,1\}, y_k \in \{0,1\} \quad \forall e\in E, k \in K. & \label{eq:integer}
\end{flalign}
In the above formulation, the constraints \eqref{eq:degree} are the degree constraints for the targets and they ensure the number of edges incident on any target is $2$. The constraints \eqref{eq:sec} are the sub-tour elimination constraints and they ensure that any feasible route for the vehicle has no sub-tours of any subset of the targets $V$. The constraints \eqref{eq:sensor} ensure that each edge $e$ that is traversed by the vehicle is covered by a subset of installed LMs to enable vehicle localization as it traverses the edge $e$. Finally, the constraints \eqref{eq:integer} are the binary restrictions on the $x_e$ and $y_k$ variables. In the next section, we present a branch-and-cut algorithm to solve the mathematical formulation to optimality.

\section{Branch-and-cut algorithm} \label{sec:bandc}
In this section, we briefly present the main ingredients of a branch-and-cut algorithm that is used to solve the two formulations presented in the previous section to optimality. The formulation can be provided to off-the-shelf commercial branch-and-cut solvers like Gurobi or CPLEX to obtain an optimal solution to the SVRP-LL. But, the formulation contains exponential number of sub-tour elimination constraints in \eqref{eq:sec} and complete enumeration of all the constraints to provide the formulation to the solver would be computationally intractable.   To address this issue, we use the following approach: we relax these constraints from the formulation, and whenever the solver obtains an integer solution feasible to this relaxed problem (or a fractional solution with integrality constraints dropped), we check if any of these constraints are violated by the feasible solution, integer or fractional. If so, we add the infeasible constraint and continue solving the original problem; we refer to this technique as a dynamic cut-generation or cutting plane algorithm. This process of adding constraints to the problem sequentially has been observed to be computationally efficient for the TSP and some of its variants \cite{Sundar2016,Sundar2015a,Sundar2016a}. The algorithms used to identify violated constraints are referred to as separation algorithms. For the sake of completeness, a detailed pseudo-code of the branch-and-cut algorithm for the SVRP-LL is given detailed. To that end, let $\bar \tau$ denote the optimal solution to an instance of SVRP-LL. \\

\noindent S\textlcsc{tep} 1 (Initialization). Set the iteration count $t \gets 1$ and the initial upper bound to the optimal solution $\bar \alpha \gets + \infty$. The initial linear sub-problem is then defined by formulation in Sec. \ref{sec:formulation} without the sub-tour elimination constraints in \eqref{eq:sec} and the binary restrictions on the variables relaxed. The initial sub-problem is solved and inserted in a list $\cal L$. 

\noindent S\textlcsc{tep} 2 (Termination check and sub-problem selection). If the list $\cal L$ is empty, then stop. Otherwise, select a sub-problem from the list $\cal L$ with the smallest objective value. 

\noindent S\textlcsc{tep} 3 (Sub-problem solution). $t \gets t + 1$. Let $\alpha$ denote the objective value of the sub-problem solution. If $\alpha \geqslant \bar \alpha$, then proceed to S\textlcsc{tep} 2. If the solution is feasible for the SVRP-LL, then set $\bar \alpha \gets \alpha$, update $\bar \tau$ and proceed to S\textlcsc{tep} 2. 

\noindent S\textlcsc{tep} 4 (Constraint separation and generation). Using the separation algorithm for the sub-tour elimination constraints, identify the violated constraints \eqref{eq:sec}. Add the violated constraints to the initial linear sub-problem and proceed to S\textlcsc{tep} 3. If no constraints are generated, then proceed to S\textlcsc{tep} 5. 

\noindent S\textlcsc{tep} 5 (Branching).
Create two sub-problems by branching on a fractional $x_e$ or $y_i$ variable. Then insert both the sub-problems in the list $\cal L$ and go to \noindent S\textlcsc{tep} 2. \\

Now, we detail the separation algorithm used in \noindent S\textlcsc{tep} 4 to identify violated sub-tour elimination constraints. To that end, let $G^* = (V^*, E^*)$ denote the support graph associated with a given fractional solution $(\bm x^*, \bm y^*)$ i.e., $V^* = T$ and $E^* := \{e \in E: x_e^* > 0\}$. Here, $\bm x$ and $\bm y$ are the vector of decision variable values in SVRP-LL. Next, we examine the connected components in $G^*$. Each connected component that does not contain all the targets in $T$ generates a violated sub-tour elimination constraint for $S = C$. If the number of connected components is one, then the most violated constraint of the form $x(\delta(S)) \geqslant 2$ can be obtained by computing the global minimum on a capacitated undirected graph $G^*$; let the cut be denoted by $(S, V^* \setminus S)$. $S$ defines a violated sub-tour elimination constraint if the value of the cut is strictly less than 2. 

\section{System architecture} \label{sec:arch}
The branch-and-cut algorithm in the previous section provides the sequence in which the vehicle should visit the targets and the locations where the LMs should be placed. The constraints ensure that the vehicle can perform localization when the LMs when placed at the specified locations. 
\begin{figure}
    	\centering
		\includegraphics[scale=0.7]{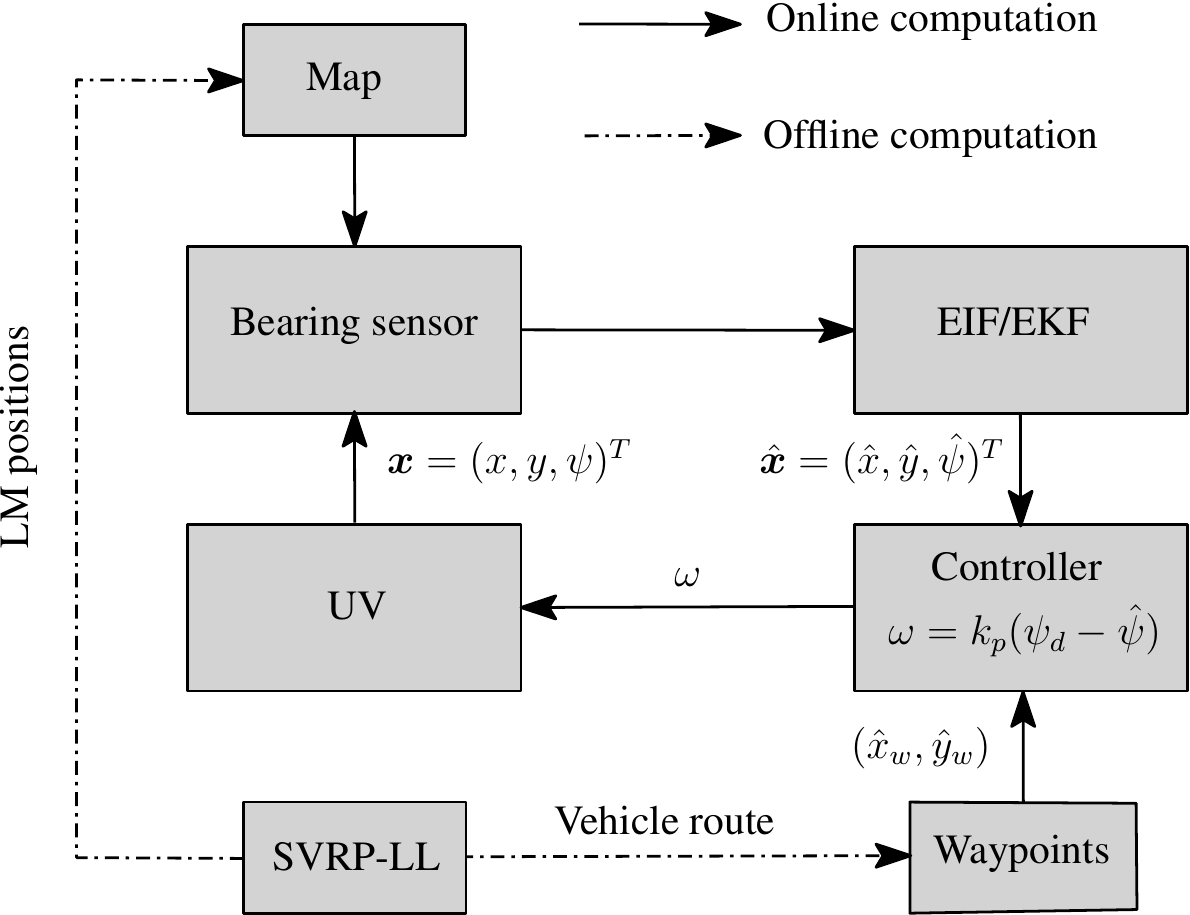}
		\caption{Block diagram showing the system architecture for estimating the states.}
		\label{fig:arch}   
\end{figure}
The Fig. \ref{fig:arch} shows the block diagram of the system architecture that is used for estimating the states using the bearing measurements from the LMs. The dashed lines denote that the computation is performed offline, which in this case is the solution to the optimization problem using the branch-and-cut algorithm. The sequence in which the UV visits the targets provides the way points for the paths. The bearing sensors are the exteroceptive sensors on the vehicle which obtain the bearing information from the LMs placed on the locations chosen by the optimization problem. This bearing measurements are in turn provided to the EIF which estimate the states of the system $\bm{\hat x}$. This is provided as input to a proportional controller that computes the corresponding $\omega$, the control input to the vehicle. The effect of choosing different values for gain of the proportional controller is detailed in the forthcoming section on computational and simulation results. 

\section{Computational and simulation results} \label{sec:results}
This section presents extensive computational and simulation results for all the algorithms developed thus far. 
All the computational experiments were performed on a MacBook Pro with a 2.9 GHz Intel Core i5 processor and 16 GB RAM using CPLEX 12.7 as a mixed-integer linear programming solver. 

\subsection{Instance generation} \label{subsec:instance}
The number of targets, $|V|$, for all the test instances was chosen from the set $\{15, 20, 25, 30\}$. For each value of $|V|$, $20$ random instances were generated. The targets were randomly placed in a $100 \times 100$ grid. As for the potential LM locations, $5 \cdot |V|$ locations on the $100 \times 100$ grid was chosen at random. In total, we had $80$ instances on which all the experiments were performed. The sensing range for the bearing sensors, $\rho_s$, was fixed at $35$ units.

\subsection{Branch-and-cut algorithm performance} \label{subsec:bandc}
The branch-and-cut algorithm with the dynamic cut-generation routine presented in Sec. \ref{sec:bandc} was implemented in C++ using the callback functionality of CPLEX. The internal cut-generation routines of CPLEX were switched off and CPLEX was used only to manage the enumeration tree in the branch-and-cut algorithm. All computation times are reported in seconds. The performance of the algorithm was tested on randomly generated test instances. The branch-and-cut algorithm to compute optimal solution to the problem is very effective in computing optimal solutions for instances with less than $30$ targets. The computation time for all the instances was less than a seconds and hence, for all practical purposes it can be converted to an online computation if $|V| \leq 30$. 

\begin{figure}
	\centering
\begin{tikzpicture}[scale=0.9]
\begin{axis}[
	x tick label style={
		/pgf/number format/1000 sep=},
	ylabel=Average values,
	xlabel=Number of targets,
	enlargelimits=0.05,
	legend style={at={(0.38,0.95)}, draw=none,
	anchor=north},
	ybar interval=0.5,
	transpose legend,
 	legend cell align=left
]
\addplot
	coordinates {(15,10.27) (20,12.21) (25,14.90) (30,20.70) (35, 0)};
\addplot
	coordinates {(15,8.89) (20,9.05) (25,9.05) (30,8.50) (35, 0)};
\legend{\# constraints \eqref{eq:sec} added, \# landmarks placed }
\end{axis}
\end{tikzpicture}
\caption{Average number of sub-tour elimination constraints in \eqref{eq:sec} and the average number of landmarks placed by the optimal solution to the SVRP-LL instances.}
\label{fig:avg}
\end{figure}
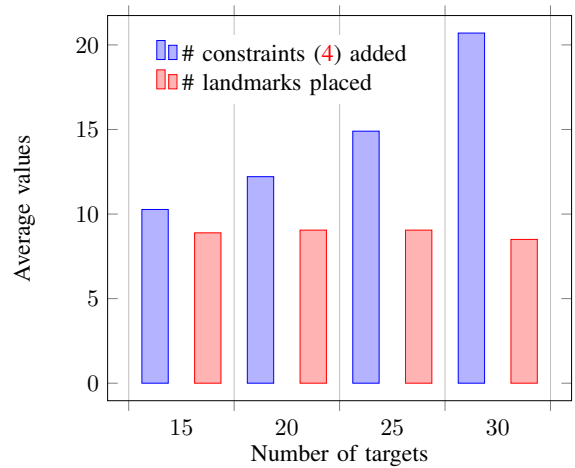

The Fig. \ref{fig:avg} shows the average number of sub-tour elimination constraints in \eqref{eq:sec} added by the dynamic cut generation procedure detailed in Sec. \ref{sec:bandc} and the average number of LMs required in the optimal solution; this indicates the effectiveness of the dynamic cut generation approach. The average number of LMs remains fairly steady as the number of targets is increased indicating that it is a function of the area of the grid.

\subsection{Simulation results} \label{subsec:simulation}
For the simulation \textit{i.e.}, online estimation algorithm using the results of the branch-and-cut algorithm, we consider $3$ cases where the route for the UV and the positions of the landmarks are known a prioi (using the branch-and-cut algorithm). 
In all cases, the vehicle is required to travel in an optimal path such that it covers all way points and has connection to at least two LMs at all time; this constraint is enforced by the formulation presented in Sec. \ref{sec:formulation}. For every run, the controller gain was chosen such that a minimum distance requirement condition to each way point was maintained. This in turn implies that for the vehicle to switch from the current way point to the next one, the vehicle needs to come in close proximity to the current way point meeting at least the minimum distance requirement for the switching to take place. For this simulation, it was considered that the vehicle can have a very high turn rate such that on reaching a way point, it can immediately point towards the next way point. The  estimated states for the vehicle were used in the way point controller instead of the true states to show that the vehicle can indeed travel optimally in a GPS-restricted environment provided that the condition for path to at least two LMs is always maintained. The simulations were ran for $3000$ iterations. For the purpose of simulation, the unit for distance and time is chosen as meters (\emph{m}) and seconds (\emph{s}), respectively, without loss of generality. An instance of the simulation has been provided below in Fig. \ref{fig:C1_R35L15_OTG} denoting the Landmarks, way points, true and estimated position of the vehicle, the associated uncertainty ($3\sigma$) bound ellipse and vehicle's path to LM(s) within sensing range.

\begin{figure}
\centering
\includegraphics[scale=0.45] {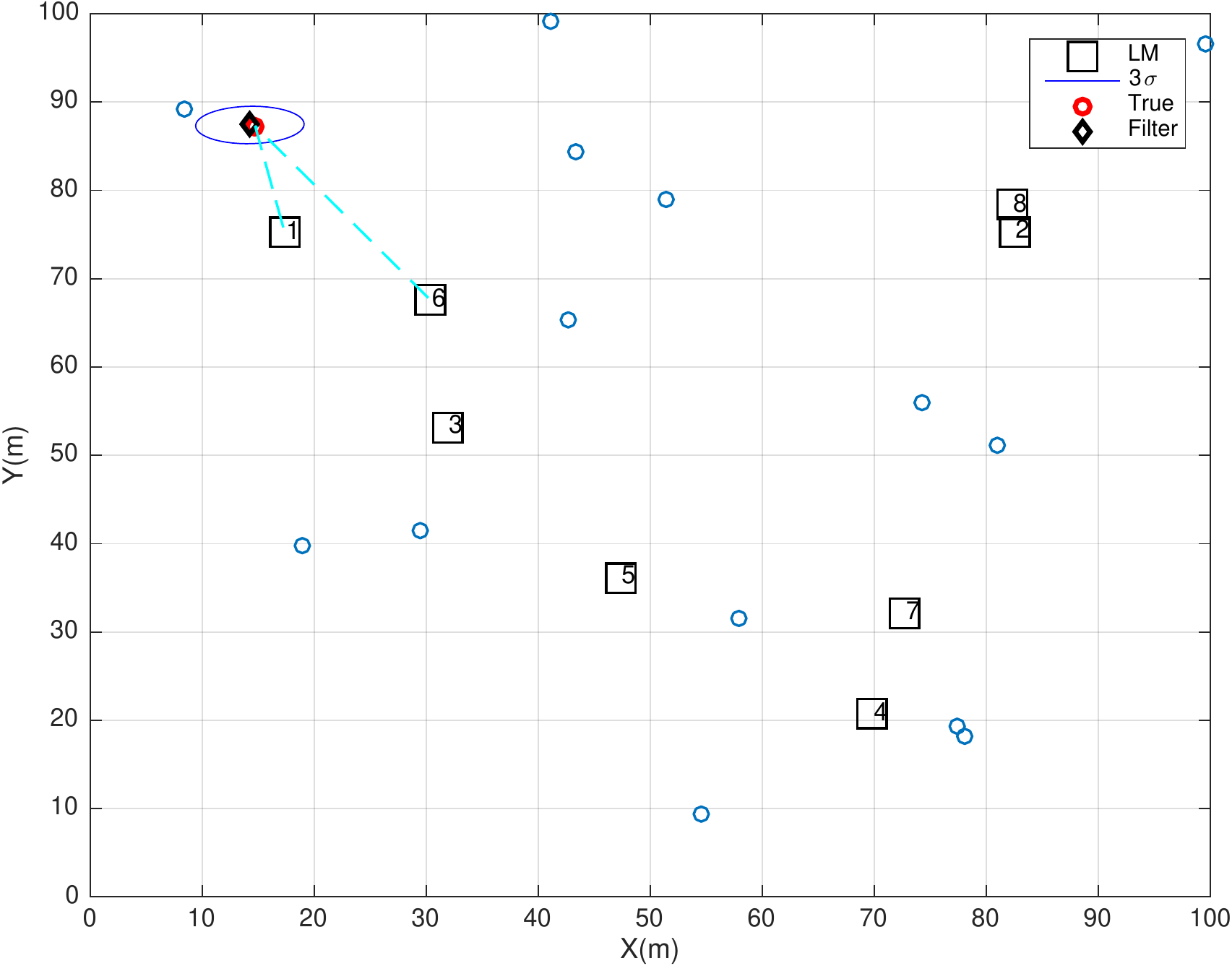}
\caption{Plot showing the vehicle motion at an arbitrary point in time during the simulation with sensing range as $35$ units for $|V| = 15$.}
\label{fig:C1_R35L15_OTG}
\end{figure}

In all cases, there were $8$ LMs and the starting position for the vehicle was chosen to be $(0,35)$, without loss of generality. The plots for each of the $4$ cases and the different conditions in which these instances were ran are provided below. There are 2 distinct categories of plots that are provided for each simulation and they contain the following information: the first set of plots show the true and estimated trajectories for the vehicle and the second set of plots show the error plots along with the $3\sigma$ bounds for each case.

\subsubsection{Case 1}	
\begin{figure}
	\centering
	\includegraphics[width = 80mm,keepaspectratio]{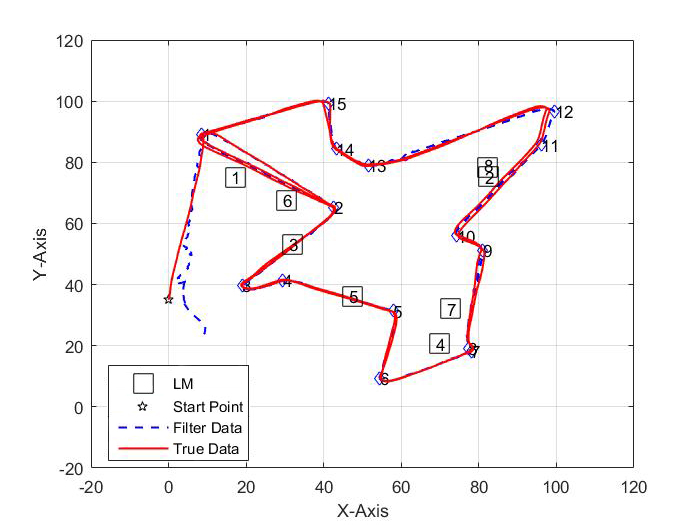}
	\caption{Plot showing trajectory of the vehicle with sensing range as $35$ units and minimum distance to WPs as $1.0$ unit (first scenario) for $|V| = 15$.}
	\label{fig:C1_R35L15_M1T}
\end{figure}

\begin{figure}
	\centering
	\includegraphics[width = 80mm,keepaspectratio]{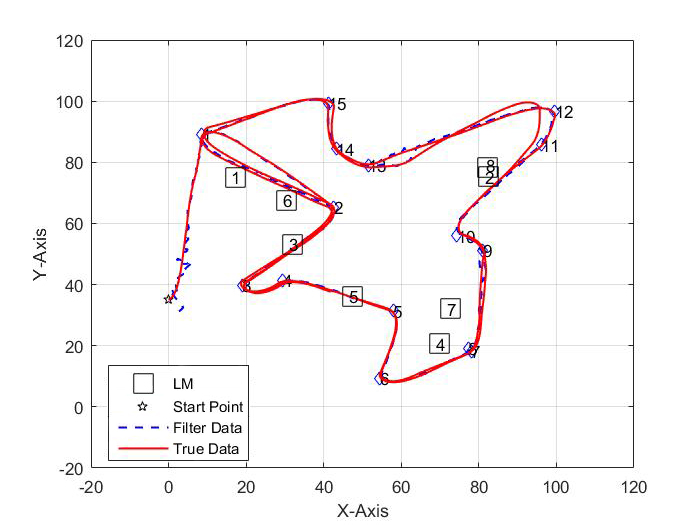}
	\caption{Plot showing trajectory of the vehicle with sensing range as $35$ units and minimum distance to WPs as $2.0$ units (second scenario) for $|V| = 15$.}
	\label{fig:C1_R35L15_M2T}
\end{figure}

\begin{figure}
	\centering
	\includegraphics[width = 80mm,keepaspectratio]{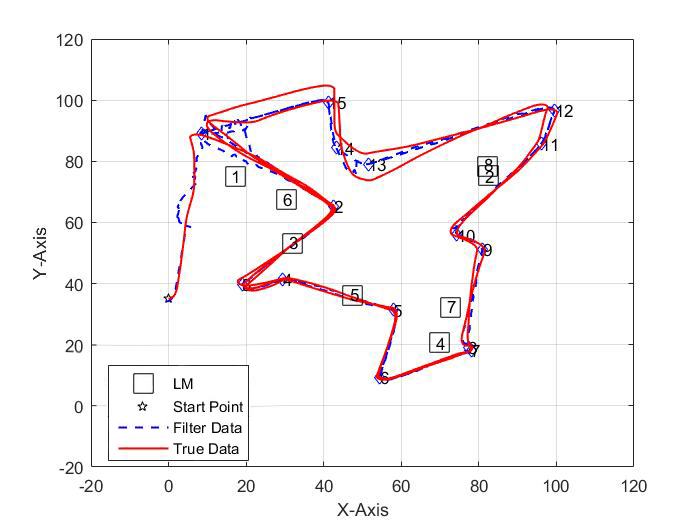}
	\caption{Plot showing trajectory of the vehicle with sensing range as $20$ units and minimum distance to WPs as $1.0$ unit (third scenario) for $|V| = 15$.}
	\label{fig:C1_R20L15_M1T}
\end{figure}

\begin{figure}
	\centering
	\includegraphics[width = 83mm,keepaspectratio]{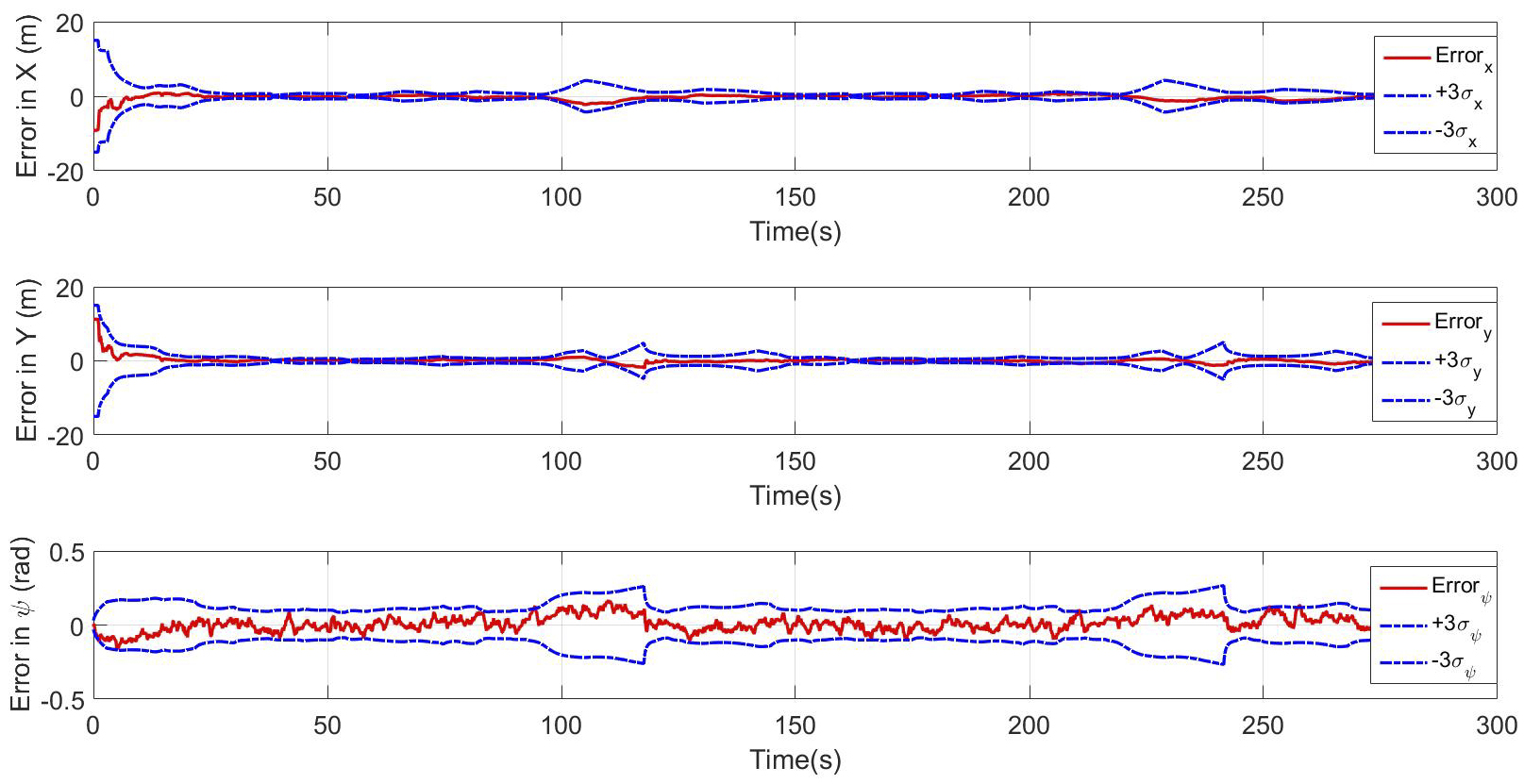}
	\caption{Plot showing the error in X direction, Y direction and heading $(\psi)$ along with their respective $3\sigma$ bounds for the first scenario with $|V| = 15$}
	\label{fig:C1_R35L15_M1E}
\end{figure}

\begin{figure}
	\centering
	\includegraphics[width = 83mm,keepaspectratio] {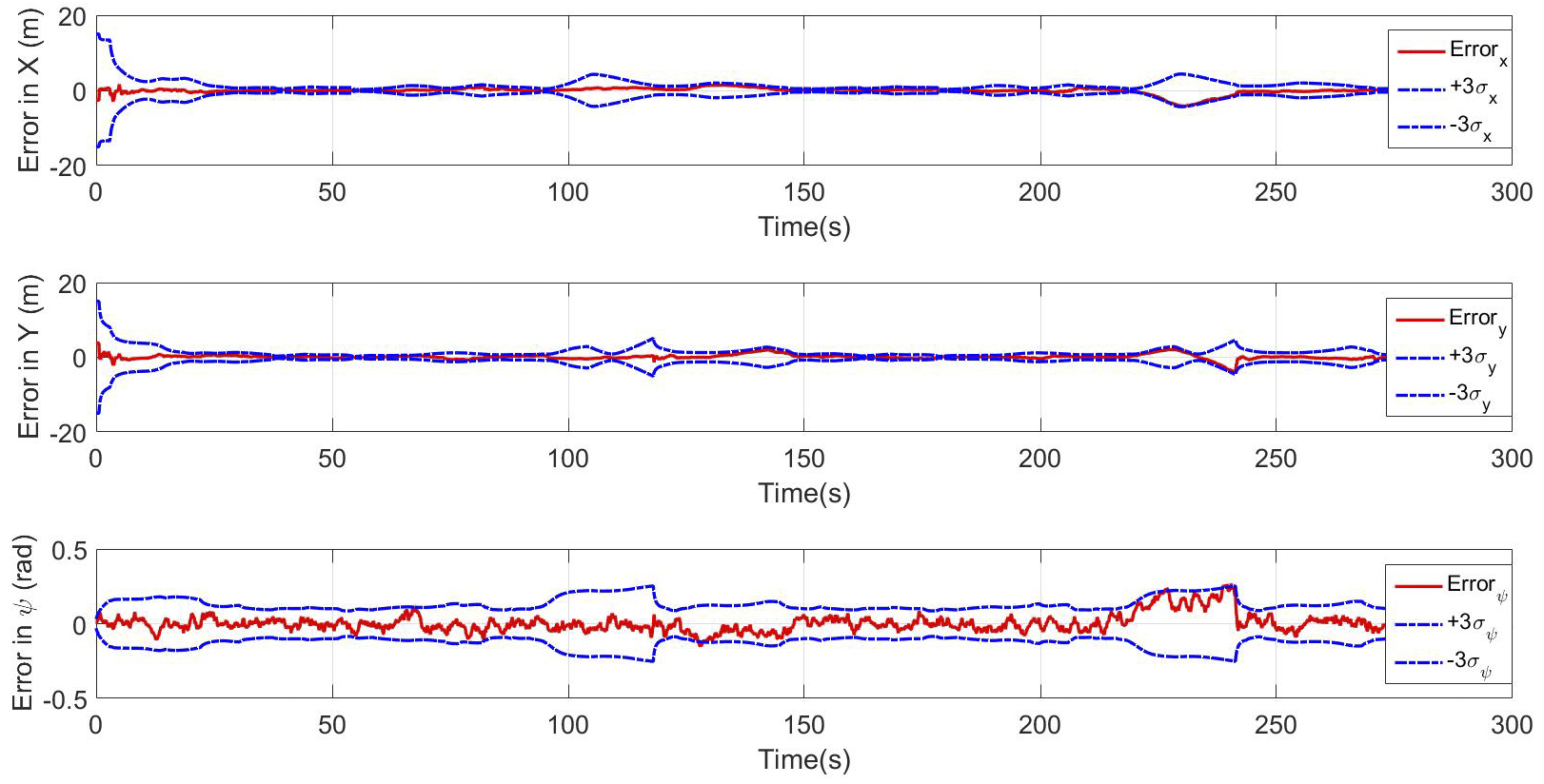}
	\caption{Plot showing the error in X direction, Y direction and heading $(\psi)$ along with their respective $3\sigma$ bounds for the second scenario with $|V| = 15$}
	\label{fig:C1_R35L15_M2E}
	\end{figure}

\begin{figure}
		\centering
		\includegraphics[width = 83mm,keepaspectratio] {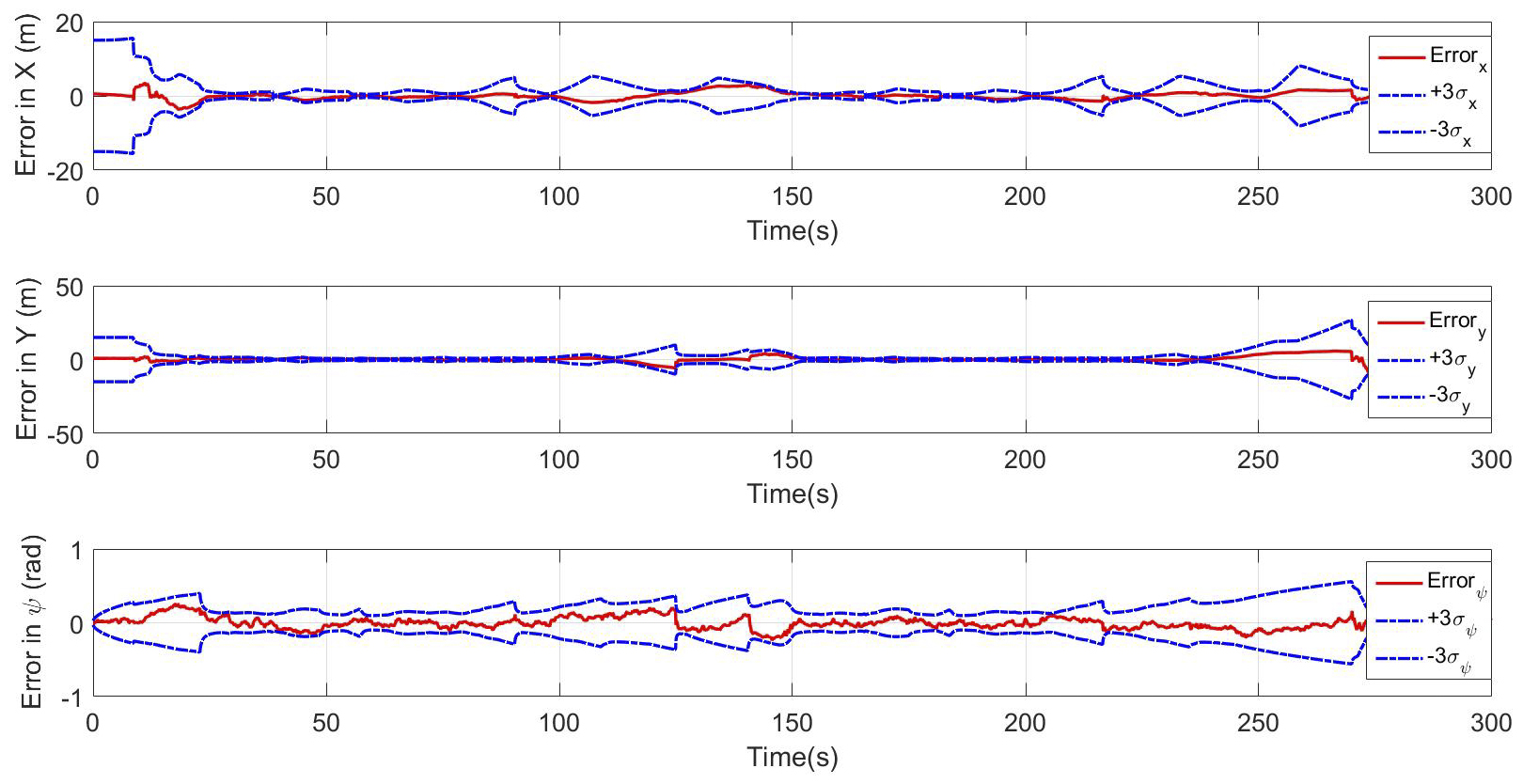}
		\caption{Plot showing the error in X direction, Y direction and heading $(\psi)$ along with their respective $3\sigma$ bounds for the third scenario with $|V| = 15$}
		\label{fig:C1_R20L15_M1E}
	\end{figure}

In this instance, $15$ way points (WPs) were provided through which the vehicle needed to route. The first scenario is such that the controller gain is set to $2.0$ and the minimum distance to WPs is $1.0$ unit. On reaching a distance of $1$ unit or closer to the WP, the vehicle can turn sharply and head towards the next WP due to high controller gain value. In the second scenario, the gain is reduced to $1.0$ and the minimum distance to WP is set at a value of $2.0$ units. A third and final scenario is provided for this particular case in which the sensing range was dropped to $20$ units from the required value of $35$ units such that the condition for path to at least 2 LMs are not always maintained. The controller gain was kept at $2.0$ and the minimum distance to WP at $1.0$ unit. The vehicle still routes through the WPs provided, but it does so with larger deviation and higher uncertainty. It can be seen from Fig. \ref{fig:C1_R20L15_M1T} that there exist a large deviation from desired path during transit from WP-15 to WP-1 and from WP-12 to WP-13 as compared to Fig. \ref{fig:C1_R35L15_M1T} and Fig. \ref{fig:C1_R35L15_M2T}. This is because only one or no LM was visible at some points in this path for a reduced sensing range. The errors and $3\sigma$ bounds are also higher in this case (see Fig. \ref{fig:C1_R35L15_M1E}, \ref{fig:C1_R20L15_M1E}, and Fig. \ref{fig:C1_R35L15_M2E}) This validates the necessity for meeting the path to at least 2 LMs condition.
		
It is evident from the graphs that the error stays within $3\sigma$ bound at all time. The relative orientation of the vehicle with respect to LMs dictates the uncertainty ellipse at any given point.

\subsubsection{Case 2}
In this instance, $20$ WPs were provided through which the vehicle needed to route. Here, the simulation was performed for 2 scenarios; one with controller gain as $2.0$ and minimum distance to WPs as $1.0$ unit and the other with controller gain as $0.4$ and minimum distance to WPs as $5.0$ units. While the first scenario is closer to ideal behavior, it requires tighter turn rate and higher vehicle agility. The vehicle gets closer to the given WPs in the first scenario than the next one in general. As a result, the second scenario produced smoother trajectory due to reduced gain and higher deviation from WPs in general, due to relaxed minimum distance requirement, than the first one. It was observed that in the second scenario, reducing controller gain resulted in requirement for a higher turning radius. Therefore, the vehicle overshot the space covered in $100 \times 100$ sq. units at times, especially when the WPs were placed very close to the edge of the square or rectangular area.

	
\begin{figure}
	\centering
	\includegraphics[width = 80mm,keepaspectratio] {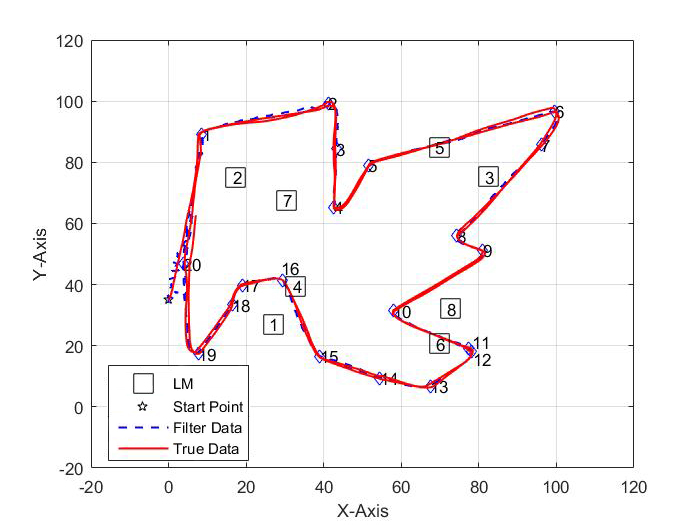}
	\caption{Plot showing trajectory of the vehicle with sensing range as $35$ units and minimum distance to WPs as $1.0$ unit (first scenario) for $|V| = 20$.}
	\label{fig:C2_R35L20_M1T}
\end{figure}

\begin{figure}
	\centering
	\includegraphics[width = 80mm,keepaspectratio] {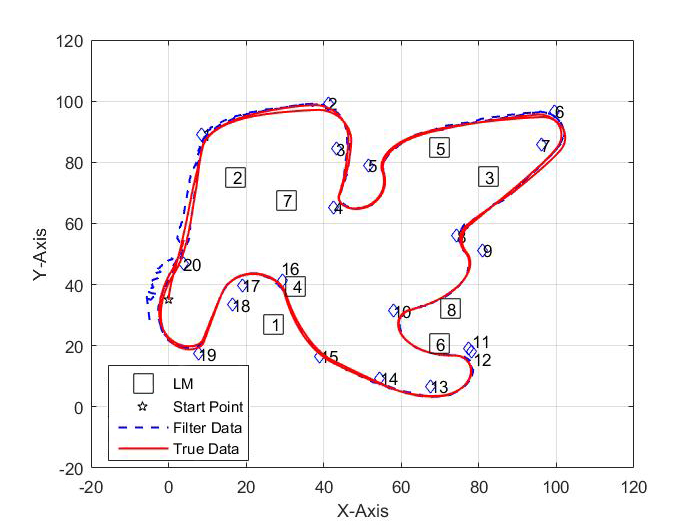}
	\caption{Plot showing trajectory of the vehicle with sensing range as $35$ units and minimum distance to WPs as $5.0$ units (second scenario) for $|V| = 20$.}
	\label{fig:C2_R35L20_M5T}
\end{figure}

Comparing Fig. \ref{fig:C2_R35L20_M1T} and Fig. \ref{fig:C2_R35L20_M5T}, it can be seen that relaxing the minimum distance requirement from 1 unit to 5 units, switching from current to next WP occurs at a much earlier time. As a result, it becomes difficult to distinguish navigation conditions, especially for closely spaced WPs as observed for WP-7, WP-9 and WP-18 particularly in Fig. \ref{fig:C2_R35L20_M5T}
	
\begin{figure}
	\centering
	\includegraphics[width = 83mm,keepaspectratio] {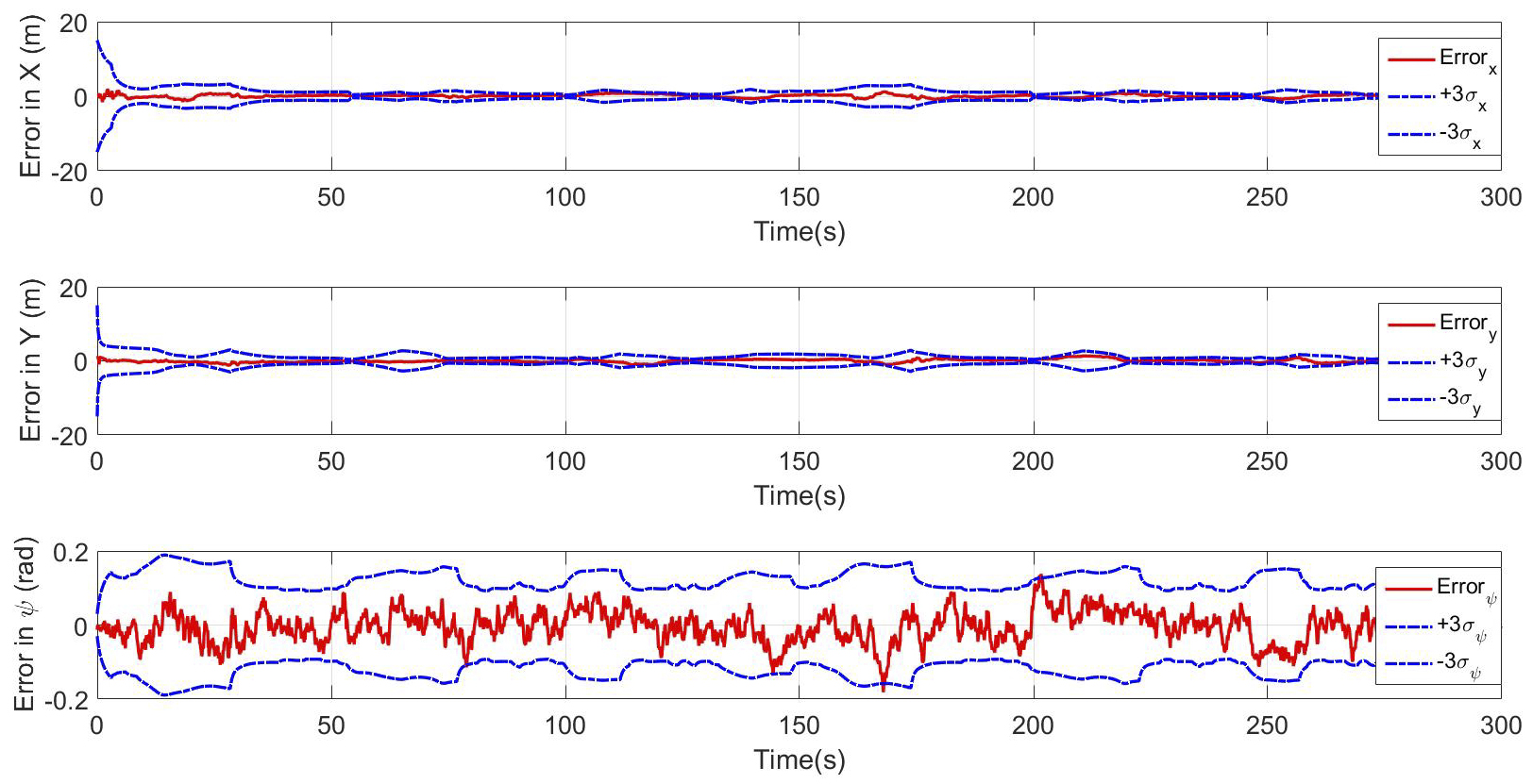}
	\caption{Plot showing the error in X direction, Y direction and heading $(\psi)$ along with their respective $3\sigma$ bounds for the first scenario with $|V| = 20$.}
	\label{fig:C2_R35L20_M1E}
\end{figure}
	
\begin{figure}
	\centering
	\includegraphics[width = 83mm,keepaspectratio] {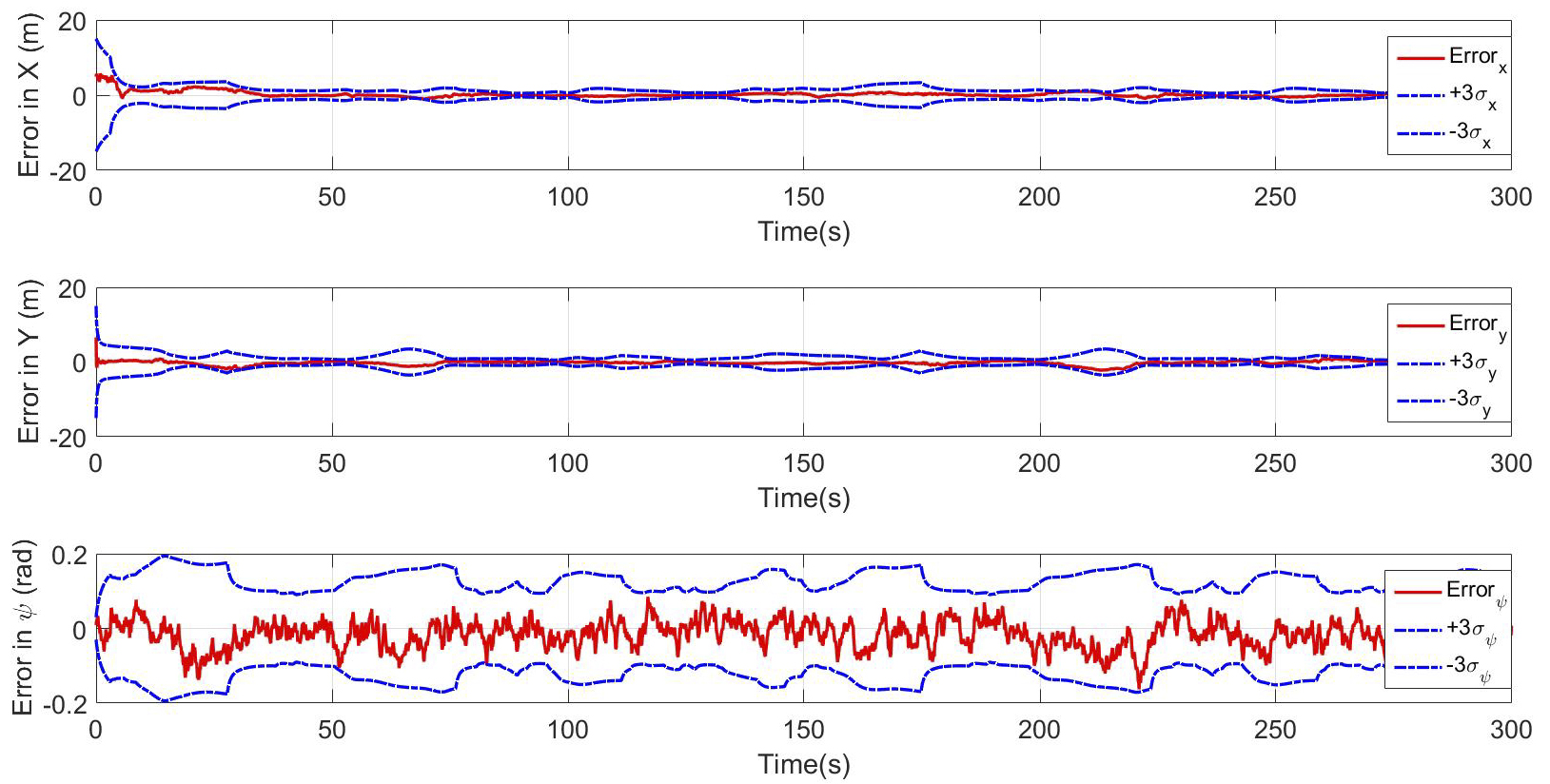}
	\caption{Plot showing the error in X direction, Y direction and heading $(\psi)$ along with their respective $3\sigma$ bounds for the second scenario with $|V| = 20$.}
	\label{fig:C2_R35L20_M5E}
\end{figure}
	
Real life scenarios would have predefined turn radius and turn rate constraints on including vehicle dynamics model. For such cases, way point navigation algorithms such as Dubins path is required to be implemented. Comparing the error plots in Fig. \ref{fig:C2_R35L20_M1E} and Fig. \ref{fig:C2_R35L20_M5E}, it is observed that the $3\sigma$ bounds are small and comparable for both scenarios. This is so because error is not dependent on the minimum distance to WPs condition. Rather, it depends on path to LMs criteria.

\subsubsection{Case 3}
In this instance, $25$ WPs were provided through which the vehicle needed to route. The simulation was performed for 2 scenarios. The first scenario had controller gain set to $2.0$ and the minimum distance to WPs as $1.0$ unit. The second scenario had controller gain tuned to $0.7$ and the minimum distance requirement to WPs as $3.0$ unit.

\begin{figure}
	\centering
	\includegraphics[width = 80mm,keepaspectratio]{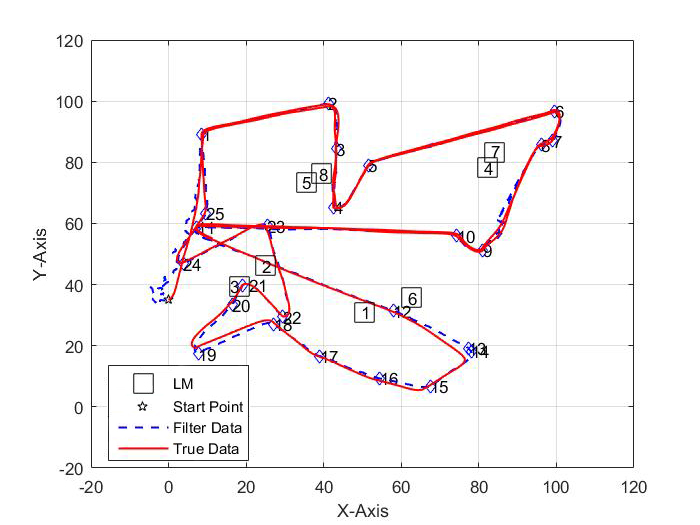}
	\caption{Plot showing trajectory of the vehicle with sensing range as 35 units and minimum distance to WPs as 1.0 unit (first scenario) for $|V| = 25$.}
\end{figure}
	
\begin{figure}
	\centering
	\includegraphics[width = 80mm,keepaspectratio]{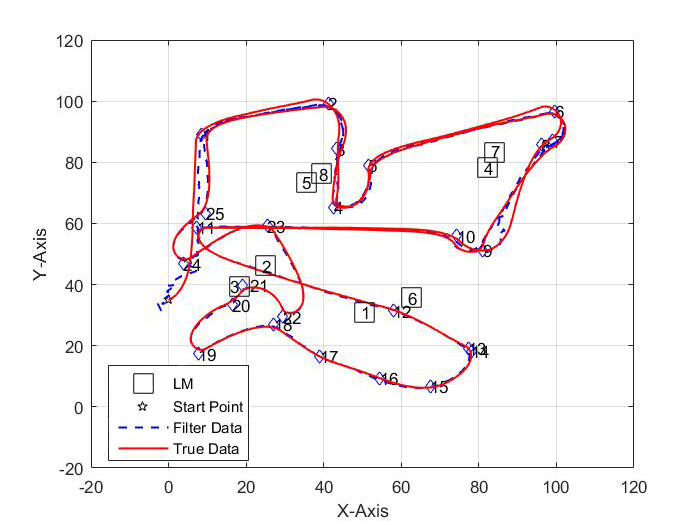}
	\caption{Plot showing trajectory of the vehicle with sensing range as 35 units and minimum distance to WPs as 3.0 units (second scenario) for $|V| = 25$.}
\end{figure}
	
\begin{figure}
	\centering
	\includegraphics[width = 83mm,keepaspectratio]{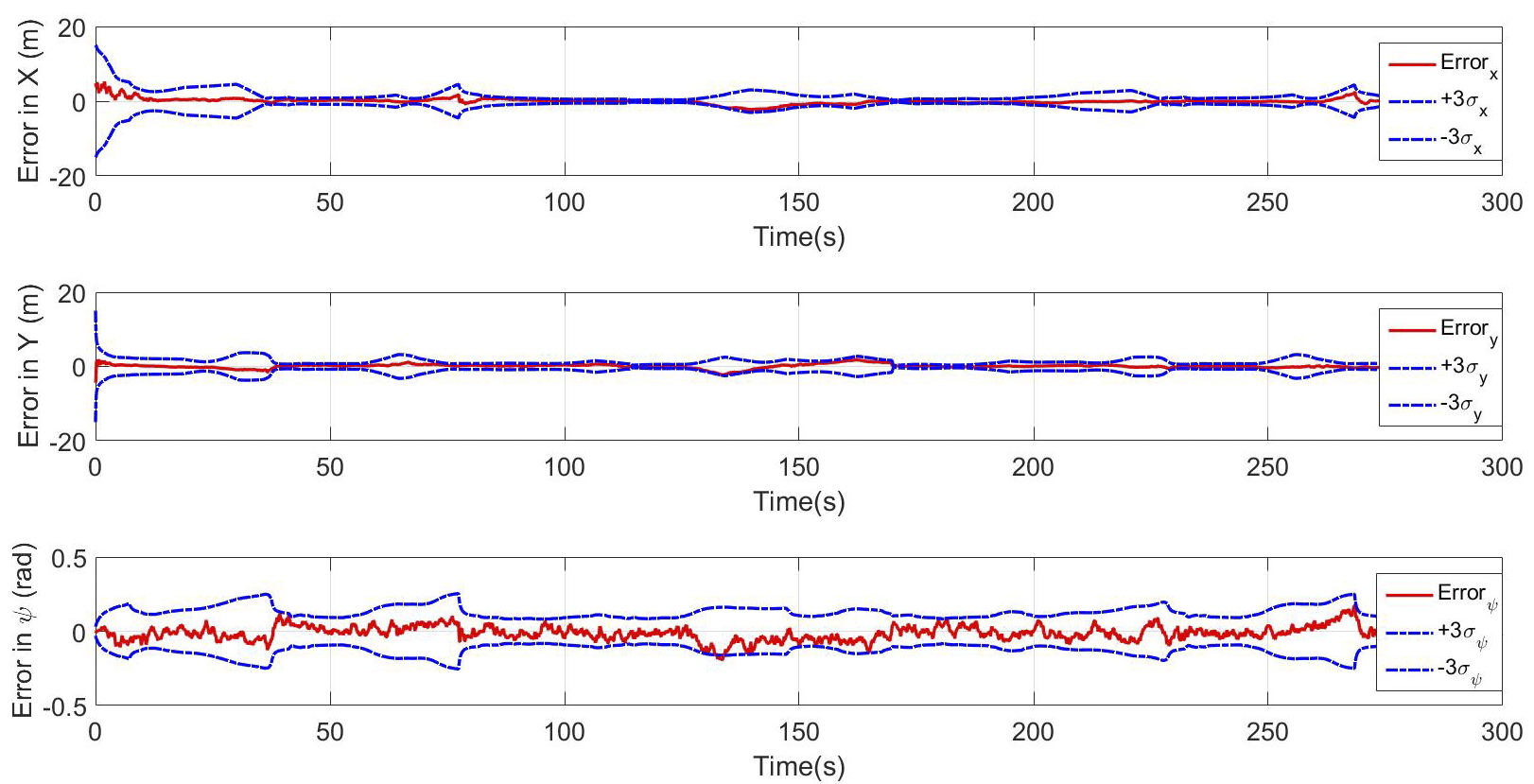}
	\caption{Plot showing the error in X direction, Y direction and heading $(\psi)$ along with their respective $3\sigma$ bounds for the first scenario with $|V| = 25$.}
\end{figure}
	
\begin{figure}
	\centering
	\includegraphics[width = 83mm,keepaspectratio]{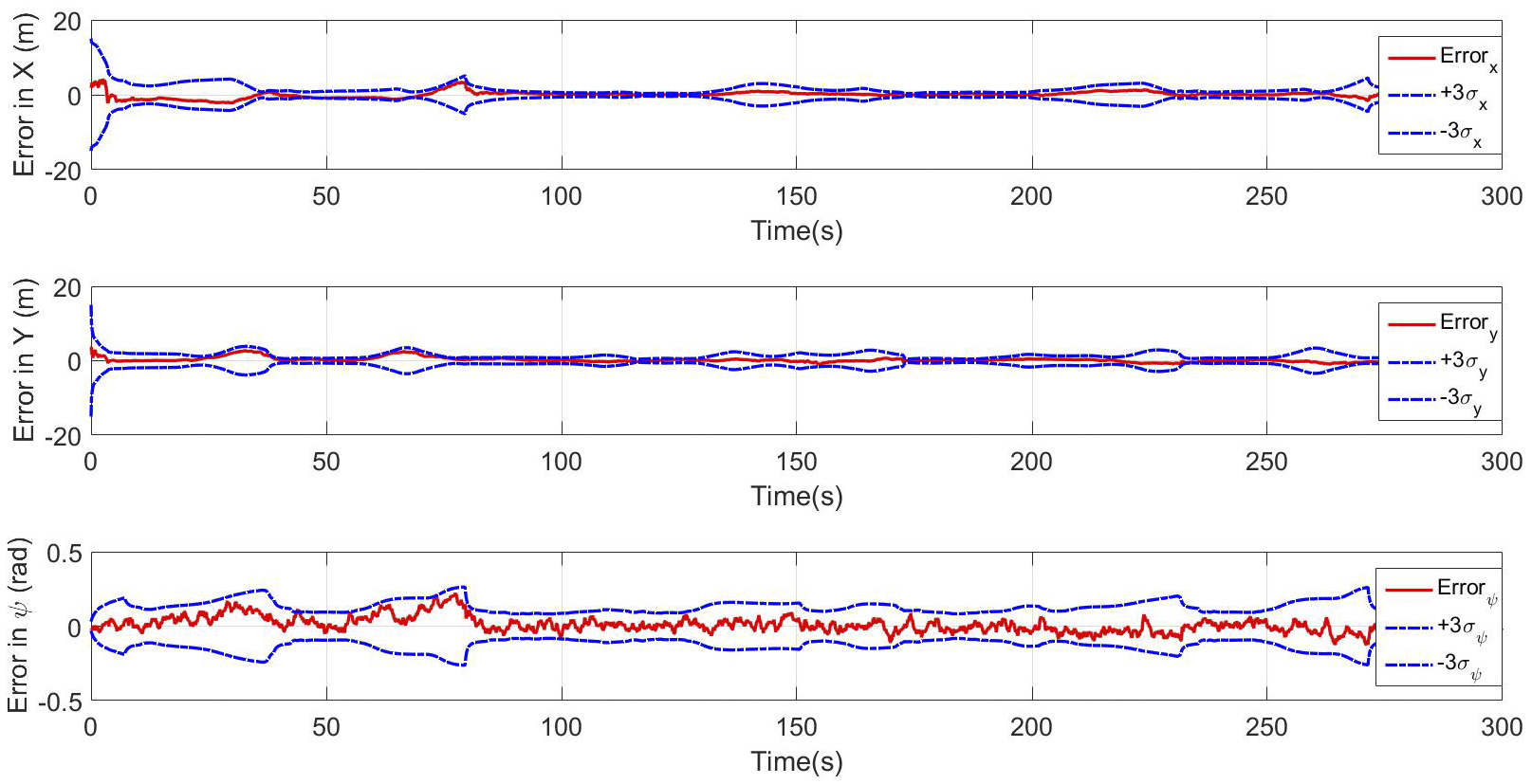}
	\caption{Plot showing the error in X direction, Y direction and heading $(\psi)$ along with their respective $3\sigma$ bounds for the second scenario with $|V| = 25$.}
\end{figure}
	
\section{Conclusion} \label{sec:conclusion}
In this paper, a systematic method to address the problem of joint routing and localization for UVs in a GPS-denied or GPS-restricted environment is presented. The optimization problem computes routes and identify the minimal set of locations where landmarks have to be placed to enable vehicle localization. This solution is combined with estimation algorithms to estimate the states of the vehicle. An efficient algorithm to compute an optimal solution to the optimization problem is presented. The proposed system architecture is tested extensively via simulation experiments. Future work can be focused on multiple vehicle versions of the problem and considering more realistic models for the sensors on the vehicles. 

\bibliographystyle{IEEEtran}
\bibliography{references}

\end{document}